\pdfoutput=1
\documentclass[11pt]{article}

\usepackage{ACL2023}

\usepackage{times}
\usepackage{latexsym}

\usepackage[T1]{fontenc}
\usepackage[utf8]{inputenc}

\usepackage{microtype}

\usepackage{inconsolata}

\usepackage{bm}
\usepackage{subfigure, graphicx}
\usepackage{multirow}
\usepackage{url}
\usepackage{booktabs}
\usepackage{enumitem}
\usepackage{amssymb} % for checkmark/tick
\usepackage{amsfonts}
\usepackage{amssymb}
\usepackage{amsmath}
\title{On the Difference of BERT-style and CLIP-style Text Encoders}

\author{
    Zhihong Chen$^{1,2*}$ \hspace{0.2cm}
    Guiming Hardy Chen$^{1,2*}$ \hspace{0.2cm}
    Shizhe Diao$^{3}$ \hspace{0.2cm} \\
    \textbf{Xiang Wan}$^{2}$ \hspace{0.2cm} 
    \textbf{Benyou Wang}$^{1,2\dag}$ \\
    $^{1}$The Chinese University of Hong Kong, Shenzhen\\
    $^{2}$Shenzhen Research Institute of Big Data\\
    $^{3}$Hong Kong University of Science and Technology\\
    \texttt{\{zhihongchen,guimingchen\}@link.cuhk.edu.cn} \hspace{0.2cm} \texttt{sdiaoaa@connect.ust.hk} \\
    \texttt{wanxiang@sribd.cn} \hspace{0.2cm} \texttt{wangbenyou@cuhk.edu.cn}
}

\begin{document}
\maketitle
\renewcommand{\thefootnote}{\fnsymbol{footnote}}
\footnotetext[1]{Equal Contribution.}
\footnotetext[2]{Corresponding authors.}
\renewcommand{\thefootnote}{\arabic{footnote}}

\begin{abstract}
Masked language modeling (MLM) has been one of the most popular pretraining recipes in natural language processing, \textit{e.g.}, BERT, one of the representative models.
Recently, contrastive language-image pretraining (CLIP) has also attracted attention, especially its vision models that achieve excellent performance on a broad range of vision tasks.
However, few studies are dedicated to studying the text encoders learned by CLIP.
In this paper, we analyze the difference between \textit{BERT-style} and \textit{CLIP-style} text encoders from three experiments: (i) general text understanding, (ii) vision-centric text understanding, and (iii) text-to-image generation.
Experimental analyses show that although CLIP-style text encoders underperform BERT-style ones for general text understanding tasks, they are equipped with a unique ability, \textit{i.e.}, \textit{synesthesia}, for the cross-modal association, which is more similar to the senses of humans.\footnote{Our code is released at \url{https://github.com/zhjohnchan/bert-clip-synesthesia}.}
\end{abstract}

% ****** table 1 ******
\begin{table*}[t]
\centering
\resizebox{\textwidth}{!}{
\begin{tabular}{@{}llrccccccccc@{}}
\toprule
\multicolumn{2}{l}{\multirow{2}{*}{Model}} & \multirow{2}{*}{Param.} & CoLA  & SST-2 & MRPC  & STS-B     & QQP   & MNLI  & QNLI  & RTE   & \multirow{2}{*}{Avg.} \\
\multicolumn{2}{l}{}                       &                         & (Mcc) & (Acc) & (F1)  & (Sp Corr) & (F1)  & (Acc) & (Acc) & (Acc) \\ \midrule
\multicolumn{12}{l}{\textit{BERT-style Text Encoders}}                                                               \\
\multicolumn{2}{l}{BERT-base}      & 110M    & 57.78 & 92.20 & 88.50 & 88.79 & 87.62 & 84.13 & 90.50 & 65.34 & 81.86 \\
\multicolumn{2}{l}{BERT-large}     & 340M    & 65.04 & 93.12 & 90.94 & 89.12 & 88.53 & 86.61 & 92.28 & 67.87 & 84.19 \\
\multicolumn{2}{l}{RoBERTa-base}   & 125M    & 58.29 & 94.50 & 91.89 & 89.96 & 88.37 & 88.00 & 92.88 & 68.23 & 84.02 \\
\multicolumn{2}{l}{RoBERTa-large}  & 355M    & \textbf{65.54} & \textbf{95.87} & \textbf{92.01} & \textbf{92.03} & \textbf{89.08} & \textbf{89.87} & \textbf{94.31} & \textbf{78.70} & \textbf{87.18} \\ \midrule
\multicolumn{12}{l}{\textit{CLIP-style Text Encoders}}                                                               \\
\multicolumn{2}{l}{ViT-B/32}       & 63M     & 30.37 & 90.48 & 72.79 & 80.52 & 84.79 & 76.28 & 81.51 & 51.99 & 71.09 \\
\multicolumn{2}{l}{ViT-B/16}       & 63M     & 27.72 & 89.45 & 76.51 & 83.80 & 85.51 & 76.90 & 83.01 & 52.71 & 71.95 \\
\multicolumn{2}{l}{ViT-L/14}       & 123M    & 30.64 & 91.51 & 82.83 & 82.26 & 85.67 & 78.38 & 82.90 & 52.71 & 73.36 \\
\multicolumn{2}{l}{ViT-L/14@336px} & 123M    & 33.85 & 91.28 & 82.57 & 82.19 & 85.42 & 77.66 & 82.92 & 53.07 & 73.62 \\ \bottomrule
\end{tabular}}
\caption{Comparisons of BERT-style text encoders and CLIP-style text encoders on the GLUE benchmark, with the number of parameters (denoted as Param.) reported. We use Matthew's correlation coefficient (Mcc) for CoLA, Spearman correlation (Sp Corr) for STS-B and F1 Score (F1) for MRPC and QQP. Top-1 accuracy (Acc) is used for the remaining datasets.}
\label{table:glue-benchmark}
\end{table*}
% ****** table 1 ******

\section{Introduction}
\textit{Text representation learning} provides a feasible solution to extract generic representations from texts, allowing models to better understand and make predictions about texts.
Normally, to perform this, a language model is pretrained on large-scale text corpora to learn text representation in a self-supervised manner, and it can be further used on downstream tasks, \textit{e.g.}, text classification and question answering~\citep{devlin2018bert}.

There are many recipes for pretraining text encoders\footnote{Similar exploration can be extended to decoder-based models as well.}~\citep{mikolov2013efficient,pennington2014glove,peters-etal-2018-elmo}.
One of the most popular ways is masked language modeling (MLM), a fill-in-the-blank task where a model uses the context words to predict masked tokens in a sequence.
For this type, BERT~\citep{devlin2018bert} and its variants~\citep{liu2019roberta,sanh2019distilbert,lan2019albert} are the representative \textit{encoder} models allowing the bidirectional perception of texts.
More recently, there has been another framework to produce text encoders, \textit{i.e.}, contrastive language-image pretraining~\cite{clip} (CLIP).
It trains image and text encoders through contrastive learning on a variety of image-text pairs, and the trained vision encoders achieve great success on vision-only tasks, especially its impressive zero-shot transfer results.
However, few studies investigated the trained text encoders.

In this work, we first conduct a pilot study in \S\ref{subsec:general-text-understanding} to benchmark BERT-style and CLIP-style text encoders in a popular natural language processing benchmark \citep{wang2018glue}. It shows that  CLIP-style text encoders significantly underperform BERT-style text encoders.

Therefore, a natural question arises: ``\textit{Are CLIP text encoders \textbf{useless byproducts} (or in which case can we make use of the CLIP text encoders?)}''. Our hypothesis for this question is that CLIP text encoder might additionally learn \textit{ visual knowledge} of textual concepts, which could be complementary to the \textit{semantic alignment} of textual concepts; the latter is well-captured by BERT text encoders. This reminds us of a phenomenon called `\textbf{synesthesia}'\footnote{Synesthesia is a  phenomenon that stimulation in a sensory or cognitive modality might unintentionally activate the perception in another sensory or cognitive modality. \textit{e.g.}, we might ``see'' shapes (vision) when listening to music (audition).}.

To validate our synesthesia hypothesis, we conduct a side-by-side comparison between \textit{BERT-style} and \textit{CLIP-style} text encoders from two aspects: 1) benchmarking them in vision-centric natural language understanding tasks in  \S\ref{subsec:vision-centric-text-understanding}; and 2) further probing the generated images from their encoded text representation in \S\ref{subsec:text-to-image-generation}.
First,  we evaluate the two types of encoders on the CxC dataset~\citep{parekh2020crisscrossed}, where the ground-truth similarity is annotated from both textual and visual perspectives;
Second, we directly generate images based on the two encoded text representations from \textit{BERT-style} and \textit{CLIP-style} text encoders, respectively. To achieve this, we train a single linear transformation layer to transfer the two types of text representations as prompts to a frozen image decoder.
Experimental analyses show that although CLIP text encoders are not comparable to BERT-style text encoders in general text understanding tasks, they have a unique ability, \textit{i.e.}, \textit{synesthesia}, to associate a text and its visual appearance.
This might inspire more studies on text encoders in the future. Our codes are available at \texttt{}

\section{Preliminaries}
In this section, we detail the BERT-style and CLIP-style text encoders in \S\ref{subsec:bert} and \S\ref{subsec:clip}, respectively.

\subsection{BERT-style Text Encoders}\label{subsec:bert}
The typical training objective of BERT-style text encoders is masked language modeling (MLM), which first masks a few tokens (usually 15\%) in a sequence and then predicts the masked tokens given the context, resembling the cloze task~\citep{taylor1953cloze}.
% To bridge the gap between pretraining and fine-tuning, BERT replaces a masked token by \texttt{[MASK]} (80\%), a random token (10\%), or an unchanged token (10\%).
Formally,  given the masked tokens $\{x_{i}\}_{i=1}^{n}$ and the masked text $T_{M}$, the model is trained to minimize
\begin{equation}
    \mathcal{L}_{MLM} = - \frac{1}{n} \sum_{i=1}^{n} \log p(x_{i} | T_{M}; \theta)
\end{equation}
where $\theta$ is the parameters of the model built upon Transformer~\cite{transformer}.
For BERT-style text encoders, pretraining data are pure texts. The commonly used corpora are BooksCorpus~\citep{Zhu_2015_ICCV} and English Wikipedia
\footnote{\url{https://dumps.wikimedia.org/}}.

\subsection{CLIP-style Text Encoders}\label{subsec:clip}
Different from MLM, CLIP learns image and text representations through image-text contrastive (ITC) pretraining.
It adopts two encoders for encoding images (denoted as $I$) and texts (denoted as $T$), named image and text encoders, respectively.
After encoding images and texts in a unified space, the similarity between an image-text pair could be obtained by using the cosine similarity function  $s(\cdot, \cdot)$.
% in a mini-batch $\mathcal{B}$ is computed.
Afterwards, given a mini-batch $\mathcal{B}$, the model is trained to minimize the following loss:
\begin{equation}
\footnotesize
\begin{split}
    \mathcal{L}_{ITC} = &-\frac{1}{|\mathcal{B}|} \sum_{i\in \mathcal{B}} \log \frac{\exp \left(s\left(I_i, T_i\right) / \tau\right)}{\sum_{k\in \mathcal{B}} \exp \left(s\left(I_i, T_k\right) / \tau\right)} \\
    &-\frac{1}{|\mathcal{B}|}\sum_{j\in \mathcal{B}} \log \frac{\exp \left(s\left(I_j, T_j\right) / \tau\right)}{\sum_{k\in \mathcal{B}} \exp \left(s\left(I_k, T_j\right) / \tau\right)}
\end{split}
\end{equation}
where $\tau$ is the temperature of the softmax function.
For CLIP-style text encoders, pretraining data are image-text datasets, \textit{e.g.}, the in-house WebImageText dataset~\cite{clip} and the publicly available LAION-5B dataset~\cite{laion5b}.
Normally, the architectures of the image encoders are CNN~\cite{cnn} or ViT~\citep{vit}, and those of the text encoders are Transformer.

\section{Pilot study: general text understanding}\label{subsec:general-text-understanding}
\subsection{Experimental settings}
For the datasets, we adopt two text classification tasks (CoLA~\citep{warstadt2019neural} and SST2~\citep{socher2013recursive}), three text similarity tasks (MRPC~\citep{dolan2005automatically}, QQP~\citep{qqp}, and STS-B~\citep{cer2017semeval}), and three inference tasks (MNLI~\citep{williams2017broad}, QNLI~\citep{rajpurkar2016squad}, and RTE~\citep{5pascal}) of GLUE.\footnote{We exclude the WNLI dataset~\citep{winograd} due to the large variance of the results on it.}
For BERT-style text encoders, we adopt the \texttt{base} and \texttt{large} versions of BERT and RoBERTa;
For CLIP-style text encoders, we adopt the text encoders of four versions of CLIP (\textit{i.e.}, ViT-B/32, ViT-B/16, ViT-L/14, and ViT-L/14@336px), where the first two text encoders share the same architecture but have different parameters (same for the last two) due to the difference of the vision branches.
We adopt the commonly used metric for each dataset.

\subsection{Empirical findings}

\paragraph{CLIP text encoders perform poorly in GLUE}
We report the results in Table~\ref{table:glue-benchmark}. 
The four CLIP-style text encoders consistently underperform BERT-style text encoders on all the datasets, where the CLIP-style text encoders achieve around 85\% of the scores of BERT-style ones on average.
Comparing among different datasets, the most significant performance gap occurs in the CoLA dataset.
The reason behind this is that CoLA is an English acceptability dataset that requires a model to identify whether a sequence of words is a grammatical English sentence.
This demonstrates that\textbf{ ITC is \textit{worse} than MLM on \underline{grammatical or syntactic properties}.}
This finding is consistent with \citet{bagofwords}, which shows that the vision-and-language models trained by ITC ignore word orders and therefore lack understanding of the compositional structure in the images and captions.

\section{CLIP-style text encoders capture visual perception for concept similarity}
\label{subsec:vision-centric-text-understanding}
\begin{table}[t]
\centering
\resizebox{0.49\textwidth}{!}{
\begin{tabular}{@{}llrcccc@{}}
\toprule
\multicolumn{2}{l}{\multirow{2}{*}{Model}} & \multirow{2}{*}{Param.} & \multicolumn{2}{c}{STS-L} & \multicolumn{2}{c}{STS-V} \\
\multicolumn{2}{l}{}                       &                         & Sp Corr      & P Corr     & Sp Corr      & P Corr     \\ \midrule
\multicolumn{7}{l}{\textit{BERT-style Text Encoders}}                                                                        \\
\multicolumn{2}{l}{BERT-base}              & 110M                    & 67.60        & 68.16      & 39.67        & 39.75      \\
\multicolumn{2}{l}{BERT-large}             & 340M                    & 69.99        & 70.61      & 42.12        & 42.32      \\
\multicolumn{2}{l}{RoBERTa-base}           & 125M                    & 67.47        & 68.13      & 39.28        & 39.54      \\
\multicolumn{2}{l}{RoBERTa-large}          & 355M                    & \textbf{70.17}        & \textbf{70.68}      & 43.54        & 43.48      \\ \midrule
\multicolumn{7}{l}{\textit{CLIP-style Text Encoders}}                                                                        \\
\multicolumn{2}{l}{ViT-B/32}               & 63M                     & 66.62        & 66.30      & 44.36        & 44.65      \\
\multicolumn{2}{l}{ViT-B/16}               & 63M                     & 67.85        & 67.70      & 44.85        & 45.17      \\
\multicolumn{2}{l}{ViT-L/14}               & 123M                    & 68.71        & 69.00      & \textbf{45.03}        & 45.37      \\
\multicolumn{2}{l}{ViT-L/14@336px}         & 123M                    & 68.72        & 68.95      & 45.02        & \textbf{45.38}      \\ \bottomrule
\end{tabular}}
\caption{Comparisons of BERT-style text encoders and CLIP-style text encoders on the language-based textual similarity (STS-L) and vision-based textual similarity (STS-V), with the number of parameters (Param.).}
\label{table:vision-centric-text-understanding}
\end{table}
% ***** table 2 ******
\paragraph{Motivation}
The aforementioned experiments show that BERT-style text encoders outperform CLIP-style text encoders on pure text tasks.
Therefore, a question is ``\textit{Is there any text task for us to testify the superior of CLIP-style text encoders?}''.
CLIP-style text encoders are trained under multi-modal settings and intuitively, they are better at \underline{associating} a text with a real-life scenario.
To find the answer, we designed a vision-centric text understanding task (described as follows).

\subsection{Experimental settings}
To design the task, we start from the CxC dataset~\citep{parekh2020crisscrossed}, an extension of the MS-COCO Caption dataset~\citep{lin2014microsoft}.
CxC contains human ratings for caption pairs, which we name STS-L (Semantic Textual Similarity from the Language perspective).
Afterwards, we construct a new dataset to conduct a vision-centric text task.
Specifically, we label each caption pair in STS-L by identifying whether it is from the same image or not (1 for the former and 0 for the latter). This new dataset is referred to as STS-V (Semantic Textual Similarity from the Vision perspective).\footnote{In MS-COCO, there are five captions for each image.}
Therefore, we have the STS-L and STS-V ratings for every caption pair.\footnote{We provide some example in Appendix~\ref{sec:vision-centric-task} for further illustration.}
We evaluate the same models as in \S\ref{subsec:general-text-understanding}.
We adopt Spearman Correlation scores (Sp Corr) and Pearson Correlation coefficient (P Corr) as the evaluation metrics.

\subsection{Empirical findings}
The results are reported in Table~\ref{table:vision-centric-text-understanding}.
We have two observations.
\paragraph{CLIP-style text encoders learn better visual perception}
CLIP-style text encoders underperform BERT-style text encoders on STS-L (same as verified in \S\ref{subsec:general-text-understanding}) but outperform them with respect to STS-V, demonstrating that CLIP-style text encoders are better at associating texts with images, more similar to human, which is consistent with the findings of \citet{bielawski2022does} and \citet{zhang2022visual}.

\paragraph{Larger BERT learns better visual perception}
Although BERT-style text encoders do not achieve promising results on STS-V, we find that large-size models (\textit{i.e.}, BERT-large and RoBERTa-large) achieve consistently better performance than their counterparts do on STS-V, probably due to the better generalization derived from their larger scale.

\section{CLIP-style text encoders capture visual perception for image generation}
\label{subsec:text-to-image-generation}
% ***** table 3 ******
\begin{table}[t]
\centering
\resizebox{0.48\textwidth}{!}{
\begin{tabular}{@{}llrccc@{}}
\toprule
\multicolumn{2}{l}{Model}          & Param.  & IS   & CLIP-S & CLIP-S (GT)             \\ \midrule
\multicolumn{5}{l}{\textit{BERT-style Text Encoders}}        & \multirow{10}{*}{25.21} \\
\multicolumn{2}{l}{BERT-base}      & 110M    & 1.01 & $22.46\pm{0.004}$  &                         \\
\multicolumn{2}{l}{BERT-large}     & 340M    & 1.01 & $22.43\pm{0.021}$  &                         \\
\multicolumn{2}{l}{RoBERTa-base}   & 125M    & 1.01 & $22.37\pm{0.049}$  &                         \\
\multicolumn{2}{l}{RoBERTa-large}  & 355M    & 1.01 & $22.41\pm0.020$  &                         \\ \cmidrule(r){1-5}
\multicolumn{5}{l}{\textit{CLIP-style Text Encoders}}        &                         \\
\multicolumn{2}{l}{ViT-B/32}       & 63M     & 1.01 & $22.57\pm0.043$  &                         \\
\multicolumn{2}{l}{ViT-B/16}       & 63M     & 1.01 & $22.59\pm0.049$  &                         \\
\multicolumn{2}{l}{ViT-L/14}       & 123M    & 1.01 & $\textbf{22.70}\pm0.032$  &                \\
\multicolumn{2}{l}{ViT-L/14@336px} & 123M    & 1.01 & $22.67\pm0.037$  &                         \\ 
\cmidrule(r){1-5}
\multicolumn{2}{l}{Random} & /    & 1.01 & $22.13\pm0.029$  &                                  \\ 
\bottomrule
\end{tabular}}
\caption{Comparison of different models regarding the IS and CLIP-S metrics on CelebAHQ, where CLIP-S (GT) denote the CLIP-S for the ground-truth pairs. We only report the standard deviation of CLIP-S. All IS have the same mean and their standard deviations are all less than 0.003, indicating that they are statistically the same.}
\label{table:text-to-image generation}
\end{table}
% ***** table 3 ******
\paragraph{Motivation}
In previous sections, we verify  the superiority of CLIP-style text encoders  by associating a text with an image on textural tasks. 
Next, we consider a question ``\textit{Why don't we directly translate a learned text representation to an image to compare their visual perception ability in a more straightforward way?}''.
To this end, we design a text-to-image generation pipeline to probe the association ability.

\subsection{Pipeline of text-to-image generation}
First, we assume the association ability is sourced from the overlap of the image representation space and the text representation space, which means that these two spaces share similar concepts.
Subsequently, under such a restricted condition, we achieve the overlap of the two spaces by introducing a single linear transformation layer to project the text space onto the image space.
In formal, given a text encoder $\mathcal{E}(\cdot)$ and an (unconditional) (generative) image decoder $D(\cdot)=p(\cdot)$, we can use the former to encode a text $T$ to text representations $\mathcal{E}(T)$ or use the latter the measure the probability $\mathcal{D}(I)=p(I)$ of a generated image $I$.
We then denote the linear transformation as $\mathcal{T}$.
Therefore, the whole probing pipeline is described as follows:
\begin{equation}
    I = \arg\max_{I} \mathcal{D}(I | \mathcal{T}(\mathcal{E}(T)))
\end{equation}
where we use the linearly transformed text representations $\mathcal{T}(\mathcal{E}(T))$ as the condition to \underline{prompt} the generation of the image $I$.
As mentioned, we only tune the parameters of the linear transformation $\mathcal{T}$, and freeze the text encoder $\mathcal{E}(\cdot)$ and the image decoder $\mathcal{D}(\cdot)$.

\subsection{Experimental settings}
For the unconditional image decoder, we adopt the VQGAN-Transformer model \citep{esser2021taming} pretrained on the images of CelebA-HQ
~\citep{karras2017progressive}. The auto-regressive Transformer can generate discrete image tokens, which can be further decoded into images through VQGAN.
We train and evaluate our model on the Multi-Modal CelebA-HQ dataset~\citep{xia2021tedigan} with 30,000 text-image pairs.
The same text encoders as in \S\ref{subsec:general-text-understanding} and \S\ref{subsec:vision-centric-text-understanding} are adopted.
We also include a random baseline, where we use random embeddings as input to the linear transformation $\mathcal{T}$.
All experiments are run 3 times with different random seeds. We use Inception Score~\cite{inceptionscore} (IS) and CLIP Score (CLIP-S) as metrics.

\subsection{Empirical findings}
\paragraph{CLIP text embedding generates better images}
We report the results in Table~\ref{table:text-to-image generation}.\footnote{We showcase the generated images in Appendix~\ref{sec:case-study}.}
The IS metric measures the realism of generated images, and it can be seen that the two types of models all achieve similar performance on the IS metric.
This owes to the fact that we start from a pretrained image decoder, which guarantees the generation of high-quality images and makes \textit{tuning a linear layer} feasible.

% t-test
\noindent\textbf{Statistical Significance}: The grouped means and standard deviations of BERT (B), CLIP (C) and Random (R) are $22.414\pm0.041$, $22.631\pm0.066$, $22.130\pm0.029$. Applying the pooled t-test between B\&C, B\&R, C\&R yields respective p-values 2.250e-9, 5.179e-8, 1.175e-8, which indicate that CLIP-S metrics for each of the three groups are significantly different from one another’s.
Therefore, after stitching text encoders and the image decoder, CLIP-style text encoders achieve higher scores on the CLIP-S metric that measures the matching of an image-text pair.
This demonstrates the effectiveness of CLIP-style text encoders on the association ability.\footnote{Both types of text encoders are not exposed to the representation space of the image decoder.}

\section{Conclusion}
Human interaction is multi-modal.
Starting from the conjecture that text encoders learned from multi-modal data have unique abilities, in this paper, we study the behavioral difference between BERT-style and CLIP-style text encoders.
We compare them from three aspects systematically: (i) general (pure) text understanding; (ii) vision-centric text understanding; and (iii) text-to-image generation.
Experimental analyses show that although CLIP-style text encoders underperform BERT-style text encoders on general text understanding tasks, they have a unique ability, \textit{i.e.}, \textit{synesthesia}, to associate a text and its visual appearance, which is more similar to human perception.

\section*{Acknowledgments}
This work is supported by Chinese Key-Area Research and Development Program of Guangdong Province (2020B0101350001), the Shenzhen Science and Technology Program (JCYJ20220818103001002), the Guangdong Provincial Key Laboratory of Big Data Computing, The Chinese University of Hong Kong, Shenzhen, Shenzhen Key Research Project (C10120230151) and Shenzhen Doctoral Startup Funding (RCBS20221008093330065). We also thank Tiannan Wang for offering valuable suggestions on the description of experiments in Section 4.

\section*{Limitations}
We highlight two limitations of our work.
First, the empirical comparisons are not conducted under fully controlled conditions, e.g., the sizes of models.
Limited by computational resources, we did not replicate different types of text encoders with the same number of parameters.
Instead, we show the results of different encoders of various sizes to reduce this effect.
Second, for the last experiment, we adopted the CLIP score to evaluate the matching between a text and its generated image.
This might raise an issue: ``\textit{Do images prompted by CLIP-style text representations guarantee a higher CLIP-S score owing to the fact that they are the same models?}''.
To answer this, we point out the reason why we adopted it.
The \textit{frozen} CLIP-style (BERT-style) text encoders are \textit{only} used to generate \textit{prompts} for image generation and the linear layer is trained to maximize the likelihood of generated images.
Yet, the CLIP score is used to measure the matching between images and texts.
Therefore, the uses of the CLIP text encoders and the CLIP score are disentangled.

\section*{Ethics Statement}
There are no ethics-related issues in this paper. The data and other related resources in this work are open-source and commonly used by many existing studies.

\bibliography{anthology,acl2023}
\bibliographystyle{acl_natbib}

\clearpage
\appendix

\section{Datasets statistics}
\label{appendix:datasets}
\paragraph{GLUE}
General Language Understanding Evaluation~\citep{wang2018glue} (GLUE) is a common benchmark for evaluating the comprehensive ability of a language model. It comprises 9 datasets of 3 different tasks. CoLA~\citep{warstadt2019neural} and SST-2~\citep{socher2013recursive} are single-sentence classification tasks. Given a sentence, a model is required to output its correct label. 
MRPC~\citep{dolan2005automatically}, STS-B~\citep{cer2017semeval}, and QQP~\citep{qqp} are sentence similarity tasks.
Given a pair of sentences, a model should output the similarity (a real value ranging from 0 to 5) of the sentence pair (STS-B) or output the correct label (same/different) of the sentence pair (MRPC and QQP).
MNLI~\citep{williams2017broad}, QNLI~\citep{rajpurkar2016squad}, RTE~\citep{1pascal,2pascal,3pascal,5pascal}, and WNLI~\citep{winograd} are natural language inference datasets.
Given a pair of sentences, a model should output a label indicating: whether a sentence entails the other (MNLI and RTE), whether they form a valid question-answer pair (QNLI), or whether they embody the same meaning (WNLI).

\paragraph{MS-COCO}
Microsoft COCO~\citep{lin2014microsoft} (MS-COCO) is a large dataset for image captioning, object detection, and object segmentation. Each image has 5 captions.

\paragraph{CxC}
Crisscrossed Captions~\citep{parekh2020crisscrossed} (CxC) is an extension of MSCOCO dataset~\citep{lin2014microsoft}. It contains 267,095 annotated pairs from 344 annotators and their 1,335,475 independent judgments. CxC consists of three sub-datasets.
For intramodality measure, Semantic Textual Similarity (STS) contains 88,054 text-text pairs, and 
Semantic Image Similarity (SIS) contains 89,486 image-image pairs.
Semantic Image Text Similarity (SITS) contains 89,555 image-text pairs for the intermodality measure.
Annotators follow a scale of 0 to 5 to rate the similarity of a given pair. Each pair is annotated multiple times by distinct annotators. The average score serves as the final score of each annotated pair.

\paragraph{Multi-Modal CelebA-HQ}
Multi-Modal CelebA-HQ~\citep{xia2021tedigan} is a large-scale human face dataset for evaluating multi-modal models. It associates each of the 30,000 high-quality images in CelebA-HQ~\citep{karras2017progressive} with 10 captions that are automatically generated using Probabilistic Context-Free Grammars (PCFG). We use the official split with 25,000/5,000 image-text pairs for training/testing, respectively.

\section{Vision-centric task}\label{sec:vision-centric-task}
Table~\ref{table:vision-centric-data} provides four samples in the STS-L and STS-V datasets. 
The first two columns \textit{Text} and \textit{STS-L} (originally named \textit{STS}) are taken from the CxC dataset. \textit{Text} stores text-text pairs, and \textit{STS-L} is the textual similarity scores provided by human annotators.
\textit{STS-V} is a column of ones (if the text-text pair describes the same image) and zeros (otherwise).
\textit{Model score} is obtained by computing the cosine similarity of a sentence pair with its embedding vectors. The embedding vectors are obtained by mean-pooling the token vectors in each sentence.
We can obtain a table similar to Table~\ref{table:vision-centric-data} for each model. We then measure the following two correlations: 
\underline{\textit{STS-L} vs. \textit{Model Score}} and \underline{\textit{STS-V} vs. \textit{Model Score}}. We use Spearman Correlation scores (Sp Corr) and Pearson Correlation coefficient (P Corr) as metrics for each correlation. Aggregating the results yields Table~\ref{table:vision-centric-text-understanding}.
With Table~\ref{table:vision-centric-text-understanding} in hand, we compare across different models within each column (the same metric). A higher score between \textit{STS-L} and \textit{Model Score} (between \textit{STS-V} and  \textit{Model Score})  indicates a better textual (visual) perception of a model.

\begin{table*}[t]
\centering
\resizebox{\textwidth}{!}{
\begin{tabular}{@{}lccc@{}}
\toprule
Text & STS-L & STS-V & Model Score \\ \midrule

A plate of breakfast food sits on a table & \multirow{2}{*}{1.24} & \multirow{2}{*}{0} & \multirow{2}{*}{0.66} \\
Chicken cordon blue and fries with a garnish & & & \\ \midrule

A kitchen counter covered with cleaning supplies and other items  & \multirow{2}{*}{2.40} & \multirow{2}{*}{0} & \multirow{2}{*}{0.60} \\
A young woman standing in the kitchen pours from a large measuring cup & & & \\ \midrule

A computer desk holding a monitor and keyboard in front of blinds  & \multirow{2}{*}{3.98} & \multirow{2}{*}{1} & \multirow{2}{*}{0.55} \\
The microwave and the television were set at the street for recycling & & & \\ \midrule

A man is flying a kite in a field  & \multirow{2}{*}{0.61} & \multirow{2}{*}{1} & \multirow{2}{*}{0.81} \\
A woman flying a kite in a blue sky & & & \\ \bottomrule
\end{tabular}}
\caption{Four samples of the dataset used in~\S\ref{subsec:vision-centric-text-understanding}.}
\label{table:vision-centric-data}
\end{table*}

\section{Case study}\label{sec:case-study}
\begin{figure*}[t]
  \centering\includegraphics[width=1\linewidth]{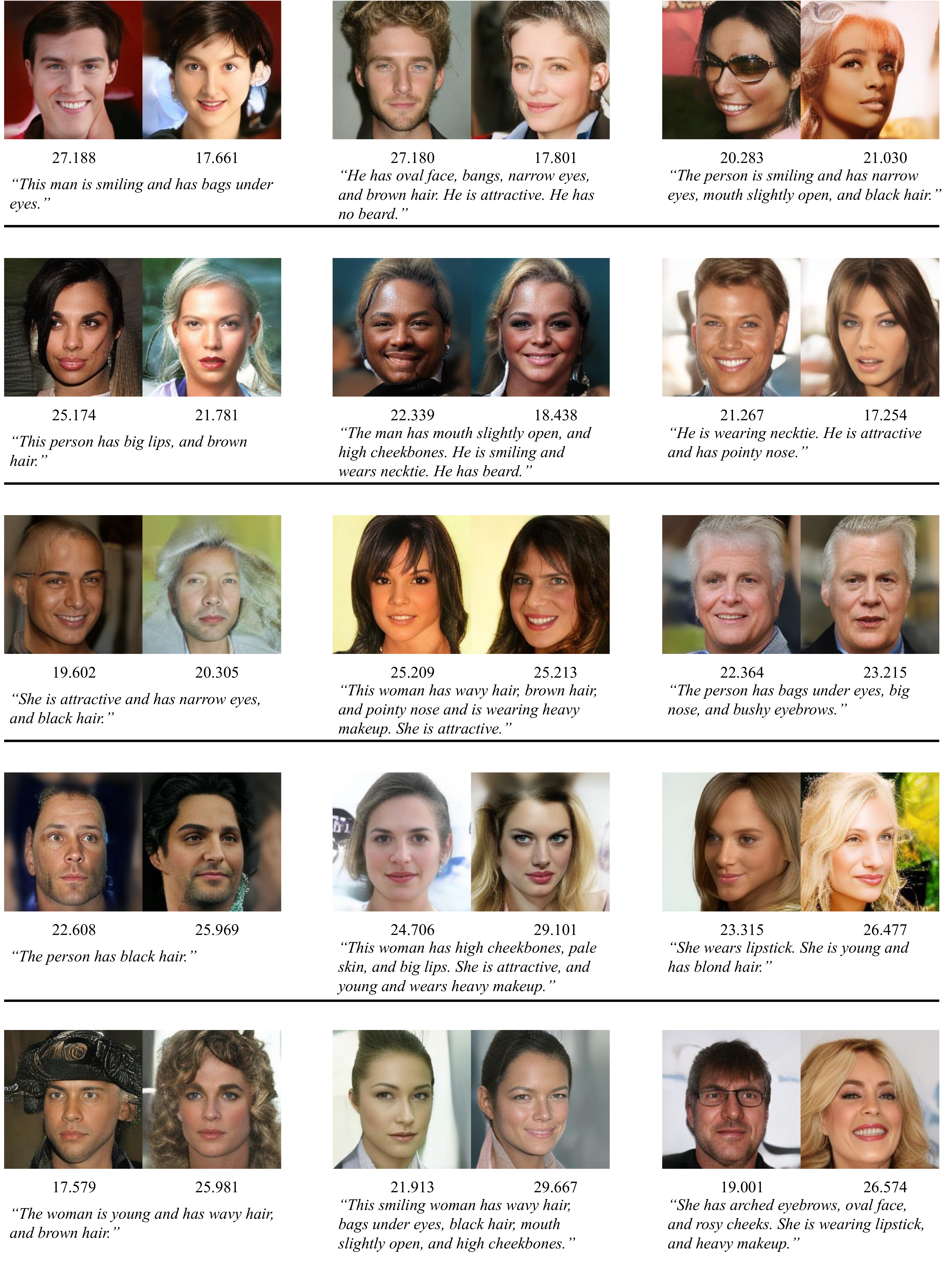}
  \caption{Case study of the image generated by the CLIP-style text encoder (left) and BERT-style text encoder (right), where the CLIP-S scores and the image captions are shown.}
  \label{fig:case-study}
\end{figure*}

We illustrate the superiority of CLIP-style text encoders in this task by showcasing some generated examples in Figure~\ref{fig:case-study}. We choose ViT-L/14 (left) and RoBERTa-large (right) as the representatives of each group with their corresponding CLIP-S score and captions.
It could be observed that the embeddings generated by the CLIP-style text encoder have a better ``sense'' of the visual world and can prompt more relevant images.
% ******** figure *********
% ******** figure *********

\end{document}